\documentclass[sigconf]{acmart}

\usepackage{booktabs}
\usepackage{multirow}
\usepackage{graphicx}
\usepackage[table,xcdraw]{xcolor}
\usepackage{balance}
\usepackage{tikz}

\newcommand{\nzk}[1]{#1}
\newcommand{\ylz}[1]{#1}

\AtBeginDocument{%
  }

\copyrightyear{2025}
\acmYear{2025}
\setcopyright{acmlicensed}
\acmConference[MM '25]{Proceedings of the 33rd ACM International Conference on Multimedia}{October 27--31, 2025}{Dublin, Ireland}
\acmBooktitle{Proceedings of the 33rd ACM International Conference on Multimedia (MM '25), October 27--31, 2025, Dublin, Ireland}
\acmDOI{10.1145/3746027.3754913}
\acmISBN{979-8-4007-2035-2/2025/10}

\begin{document}

\title{Self-Supervised Anatomical Consistency Learning for Vision-Grounded Medical Report Generation}


\author{Longzhen Yang}
\email{yanglongzhen@tongji.edu.cn}
\orcid{0000-0002-5791-145X}
\affiliation{%
  \institution{Tongji University}
  \city{Shanghai}
  \country{China}
}
\author{Zhangkai Ni}
\email{zkni@tongji.edu.cn}
\orcid{0000-0003-3682-6288}
\affiliation{%
  \institution{Tongji University}
  \city{Shanghai}
  \country{China}
}
\author{Ying Wen}
\email{ywen@cs.ecnu.edu.cn}
\orcid{0000-0002-6974-5110}
\affiliation{%
  \institution{East China Normal University}
  \city{Shanghai}
  \country{China}
}
\author{Yihang Liu}
\authornote{Corresponding authors.}
\email{211131@tongji.edu.cn}
\orcid{0000-0003-4257-2528}
\affiliation{%
  \institution{Tongji University}
  \city{Shanghai}
  \country{China}
}
\author{Lianghua He}
\authornotemark[1]
\email{helianghua@tongji.edu.cn}
\orcid{0000-0002-5250-170X}
\affiliation{%
  \institution{Tongji University}
  \city{Shanghai}
  \country{China}
}
\affiliation{%
  \institution{Shanghai Eye Disease Prevention and Treatment Center}
  \city{Shanghai}
  \country{China}
}
\author{Heng Tao Shen}
\email{shenhengtao@hotmail.com}
\orcid{0000-0002-2999-2088}
\affiliation{%
  \institution{Tongji University}
  \city{Shanghai}
  \country{China}
}

\renewcommand{\shortauthors}{Longzhen Yang et al.}

\begin{abstract}
Vision-grounded medical report generation aims to produce clinically accurate descriptions of medical images, anchored in explicit visual evidence to improve interpretability and facilitate integration into clinical workflows. However, existing methods often rely on separately trained detection modules that require extensive expert annotations, introducing high labeling costs and limiting generalizability due to pathology distribution bias across datasets. To address these challenges, we propose \textbf{S}elf-\textbf{S}upervised \textbf{A}natomical \textbf{C}onsistency \textbf{L}earning (\textbf{SS-ACL})—a novel and annotation-free framework that aligns generated reports with corresponding anatomical regions using simple textual prompts. SS-ACL constructs a hierarchical anatomical graph inspired by the invariant top-down inclusion structure of human anatomy, organizing entities by spatial location. It recursively reconstructs fine-grained anatomical regions to enforce intra-sample spatial alignment, inherently guiding attention maps toward visually relevant areas prompted by text. To further enhance inter-sample semantic alignment for abnormality recognition, SS-ACL introduces a region-level contrastive learning based on anatomical consistency. These aligned embeddings serve as priors for report generation, enabling attention maps to provide interpretable visual evidence. Extensive experiments demonstrate that SS-ACL, without relying on expert annotations, (i) generates accurate and visually grounded reports—outperforming state-of-the-art methods by \textbf{10}\% in lexical accuracy and \textbf{25}\% in clinical efficacy, and (ii) achieves competitive performance on various downstream visual tasks, surpassing current leading visual foundation models by \textbf{8}\% in zero-shot visual grounding. Our code is available at \href{https://github.com/kaelsunkiller/ssacl}{\textit{https://github.com/kaelsunkiller/ssacl}}.

\end{abstract}




\begin{teaserfigure}
  \includegraphics[width=\textwidth]{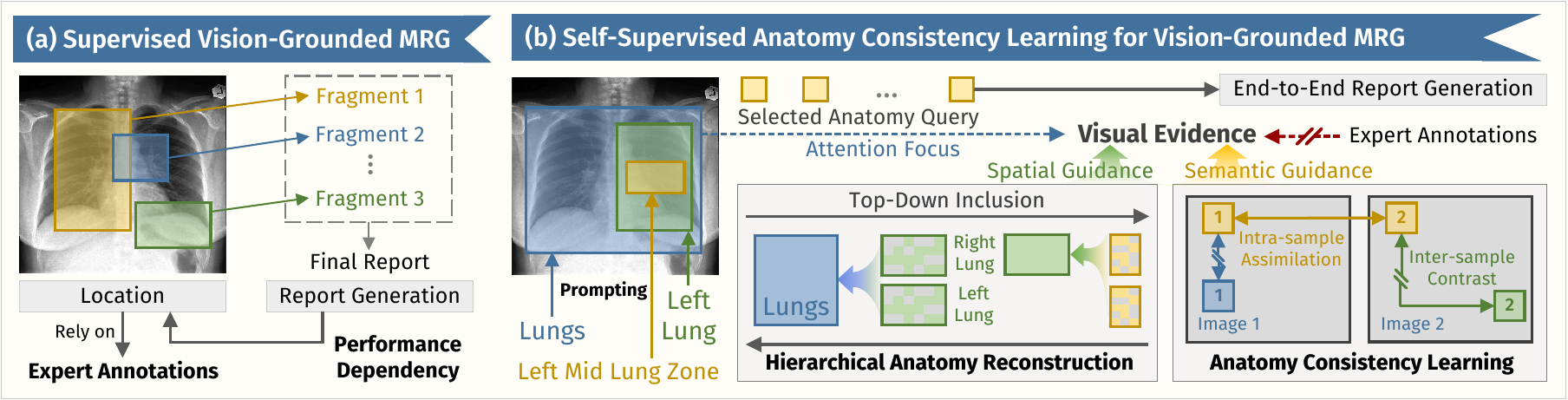}
  \vspace{-5mm}
  \caption{
  \nzk{Comparison of prior methods (a) and our SS-ACL (b):
    (a) illustrates supervised vision-grounded MRG.  
    (b) shows our SS-ACL, a self-supervised novel annotation-free framework that aligns generated reports with corresponding anatomical regions using simple textual prompts.
    }
  }
  \Description{This figure illustrates the comparison between conventional supervised vision-grounded medical report generation and our self-supervised anatomical consistency learning for vision-grounded MRG. Conventional methods locate abnormalities before report generation, which separates the training of localization and generation. This two-stage paradigm relies heavily on vast amount of expert annotations, leading to performance dependency and limited generalizability due to the pathology distribution biases between datasets. On the contrary, our SS-ACL utilizes the inherent attention machenism along with simple textual prompts to provide effective visual evidence, achieving end-to-end generation without explicit expert annotations.}
  \label{fig:teaser}
\end{teaserfigure}


\maketitle

\section{Introduction}


Automatic vision-grounded Medical Report Generation (MRG) has emerged as a promising tool to assist clinicians in the time-intensive diagnostic process. 
The goal is to produce detailed textual descriptions for each relevant anatomical region, accompanied by precise visual evidence~\cite{chen2020r2gen, tanida2023rgrg}. 
\nzk{The effectiveness of MRG largely depends on the quality of visual representations—particularly their interpretability—which is often assessed via visual grounding tasks such as detection~\cite{zhou2022refers, zhou2023mrm}, segmentation~\cite{huang2024maco, gu2024comg}, and phrase grounding~\cite{huang2021gloria, huang2024maco}.
Leveraging progress in supervised visual grounding~\cite{ronneberger2015unet, ren2016fasterrcnn}, most existing MRG frameworks adopt a two-stage pipeline (Figure~\ref{fig:teaser}(a)): initially detecting or segmenting abnormal regions, followed by generating region-specific textual descriptions conditioned on localized features~\cite{tanida2023rgrg, gu2024comg}. }
However, this non-end-to-end design often relies on heterogeneous supervision signals—such as bounding box annotations~\cite{wu2021chestimagenome} or knowledge graphs~\cite{jain2021radgraph}—for visual grounding and report generation separately.
\nzk{In practical clinical applications, these requirements significantly increase deployment complexity and limit scalability, as they are sensitive to the distributional variability of pathological patterns across datasets.}

In contrast to traditional two-stage approaches, recent advances in Medical Visual Pretraining (MVP) have shown promising capabilities in zero-shot phrase grounding through Self-Supervised Learning (SSL) techniques, such as visual-language alignment~\cite{huang2021gloria} and masked image reconstruction~\cite{huang2024maco}. However, incorporating these capabilities into end-to-end report generation remains difficult, largely due to the absence of ground-truth reports at inference. Moreover, MVP models often yield only global attention to abnormalities, whereas clinical reporting demands fine-grained, case-specific descriptions of both normal and abnormal regions~\cite{li2023dcl}.
\nzk{This gap highlights the need for a more granular and anatomically consistent approach that can operate without expert supervision while preserving semantic precision.}

\nzk{To tackle these challenges, we propose \textbf{S}elf-\textbf{S}upervised \textbf{A}natom\-ical \textbf{C}onsistency \textbf{L}earning (\textbf{SS-ACL})—a novel framework that enables robust, annotation-free visual grounding by exploiting the inherent hierarchical structure of human anatomy. 
Specifically, SS-ACL builds a \textbf{hierarchical anatomy graph} based on the consistent top-down inclusion relationships among anatomical structures, where each child node represents a subregion of its parent. 
As shown in Figure~\ref{fig:teaser}(b), this graph guides the visual encoder using text-prompted anatomical entities to produce region-specific query embeddings. 
To ensure spatial alignment, we introduce a \textbf{hierarchical reconstruction strategy}: the root query reconstructs raw image pixels from a masked image, while each parent query recursively restores its region from the masked outputs of its children. 
This recursive mechanism preserves anatomical structure and embeds restorative pixel-level cues into the queries, aligning attention maps with their corresponding regions.
To further enhance semantic alignment, we introduce \textbf{anatomical consistency learning}, a region-level contrastive learning approach that uses the inherent consistency of anatomical regions across images~\cite{haghighi2022dira, gao2024anatomy}. 
This strategy pulls together embeddings of the same anatomical entity from different images (inter-sample assimilation) and pushes apart embeddings of different regions within the same image (intra-sample contrast). 
This ensures that the learned queries are both distinctive and semantically meaningful. 
Ultimately, SS-ACL enables anatomy-aware query embeddings to act as a unified and interpretable interface for both report generation and various downstream visual tasks.
The main contribution can be summarized as:}
\begin{itemize}
\item We propose Self-Supervised Anatomical Consistency Learning (SS-ACL), an annotation-free and interpretable framework that \nzk{unifies} visual grounding and accurate report generation within an end-to-end paradigm.

\item We introduce hierarchical anatomy reconstruction and anatomical consistency learning, leveraging simple textual prompts to guide report generation that is spatially and semantically aligned—without expert annotations. 

\item \nzk{SS-ACL achieves notable gains in both visual generalization and report quality, improving lexical accuracy by 10\% and clinical efficacy by 25\%. It also generalizes well to diverse downstream vision tasks, demonstrating strong potential for broader clinical deployment.}
\end{itemize}

\begin{figure*}
    \centering
    \includegraphics[width=\textwidth]{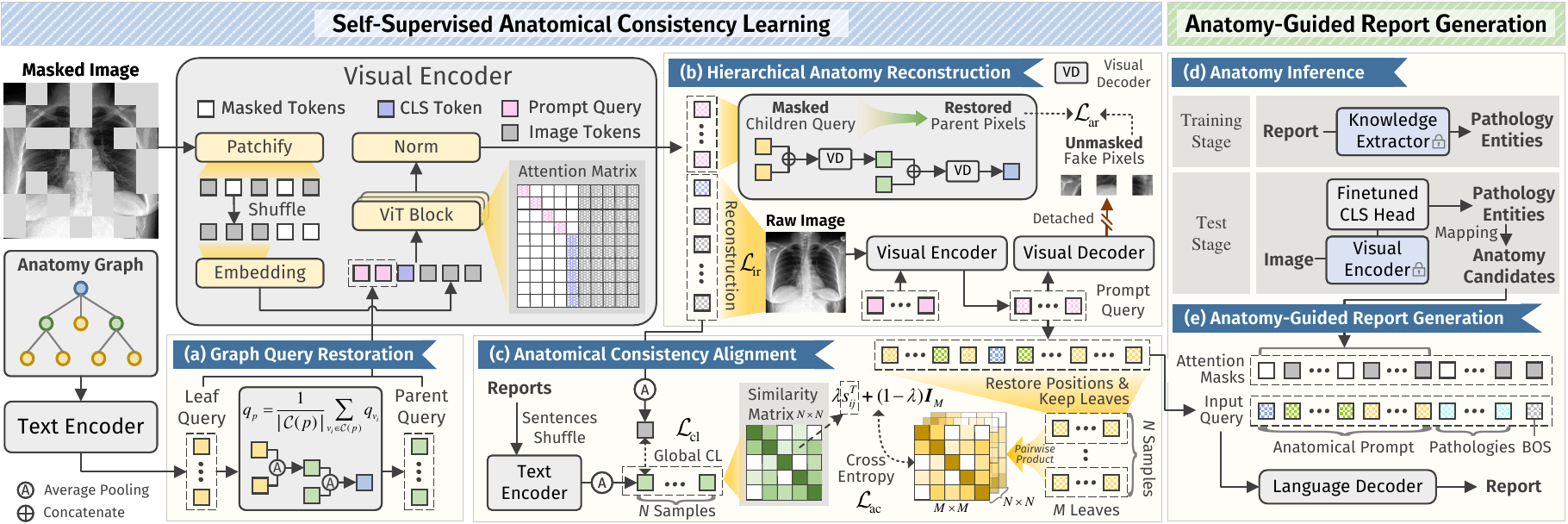}
    \vspace{-5mm}
    \caption{Overview of the proposed SS-ACL. Left: the self-supervised anatomical consistency learning for annotation-free and grounded visual encoding. Right: the subsequent report generation under the guidance of generated anatomical representations.}
    \Description{}
    \label{fig:model}
\end{figure*}

\section{Related Works}
\label{sec:related_works}

\noindent\textbf{General MRG}. Medical report generation evolved from image captioning~\cite{chen2020r2gen, chen2021r2gencmn, wang2022msat}, but general models struggle with medical-specific knowledge gaps. To address this, studies introduced domain-specific techniques such as knowledge graphs~\cite{yang2022knowledgematters, li2023dcl}, expert models~\cite{wang2023metransformer, bu2024ekagen}, and anatomical priors~\cite{tanida2023rgrg, gu2024comg}. The rise of large language models (LLMs) has further advanced MRG~\cite{wang2023r2gengpt, liu2024bllm}, though achieving synergy between visual generalizability and lexical accuracy remains challenging, limiting clinical applicability.

\noindent\textbf{Vision-Grounded MRG}. Region grounding from captioning~\cite{anderson2018butd} has been adapted to MRG through methods like RGRG~\cite{tanida2023rgrg} and COMG~\cite{gu2024comg}, which generate sentences from detected anatomical areas. However, these two-stage pipelines suffer from limited coherence and detector dependence. Many approaches also neglect anatomical hierarchies~\cite{jain2021radgraph, tanida2023rgrg, liang2024dcg}, leading to misalignment. We address this with a self-supervised framework that builds a hierarchical anatomy graph and aligns report segments accordingly, enabling more coherent, end-to-end generation.

\noindent\textbf{Medical Visual Pre-training}. MVP learns discriminative visual features for tasks like classification, segmentation, and detection. Recent methods use multimodal contrastive learning to align visual and textual representations~\cite{huang2021gloria, zhou2022refers, wu2023medklip, li2024mlip}, and reconstruction from partial information further boosts performance~\cite{zhou2023mrm, huang2024maco}. Inspired by this, MRG applies contrastive learning to guide language generation via semantically aligned visual cues~\cite{yan2021wcl, li2023dcl, wu2023mrcl, zhang2024tgr}. Yet, integrating discriminative and generative capabilities remains difficult due to the challenge of aligning visual features with the flexibility needed for accurate report generation.

\section{Methodology}

\subsection{Motivation}
\label{sec:motivation}
\ylz{When generating fine-grained descriptions for medical images with visual evidence, we intend to address the following two issues:}
\begin{itemize}
    \item \ylz{\textbf{Limited Generalization.} Current data-driven methods detect the salient regions for subsequent generation, which may provide the connection between images and texts. However, this connection is potentially biased by the pathology distribution in annotations, which limits the inherent interpretability and generalization.}
    \item \ylz{\textbf{Limited Effectiveness.} Although using self-supervised learning can improve the global visual grounding capacity without annotations, previous works do not fully consider the inherent anatomical relations for fine-grained alignment, hence reducing the effectiveness and performance improvement.}
\end{itemize}



\ylz{Our core goal is to develop an end-to-end medical report generation model with the capacity of \textbf{fine-grained visual grounding} and achieve the \textbf{synergy} of \textbf{high-quality} report generation and \textbf{generalized} visual representation learning without utilizing auxiliary expert annotations. In the following, we introduce the overall framework of our SS-ACL and detail our contributions.

\subsection{Overview}
The overall framework of SS-ACL is illustrated in Figure~\ref{fig:model}. We begin by encoding the leaf nodes of a predefined hierarchical anatomical graph using a Transformer-based text encoder. Parent nodes are then recursively restored via average pooling, forming the complete graph representation.

The resulting graph embeddings are sequentialized as a prompt query and fused with the masked input image using a standard Vision Transformer~\cite{dosovitskiy2020vit}. The resulting image tokens are used in two reconstruction tasks: i) recovering raw image pixels through an isomorphic visual decoder, and ii) reconstructing each fine-grained anatomical region by decoding associated child queries independently fused with image tokens.

We then align visual and textual representations at both global and local levels. Global embeddings are aligned through a contrastive learning scheme~\cite{wang2022medclip}, while local anatomical queries are aligned spatially under anatomical consistency constraints.

Finally, report generation is performed by feeding the anatomy queries and predicted pathology entities into a language decoder, with negative positions masked to guide accurate text generation.}

\subsection{Anatomy Graph Query Construction}
\label{sec:method:1:agc}

\begin{figure}
    \centering
    \includegraphics[width=\linewidth]{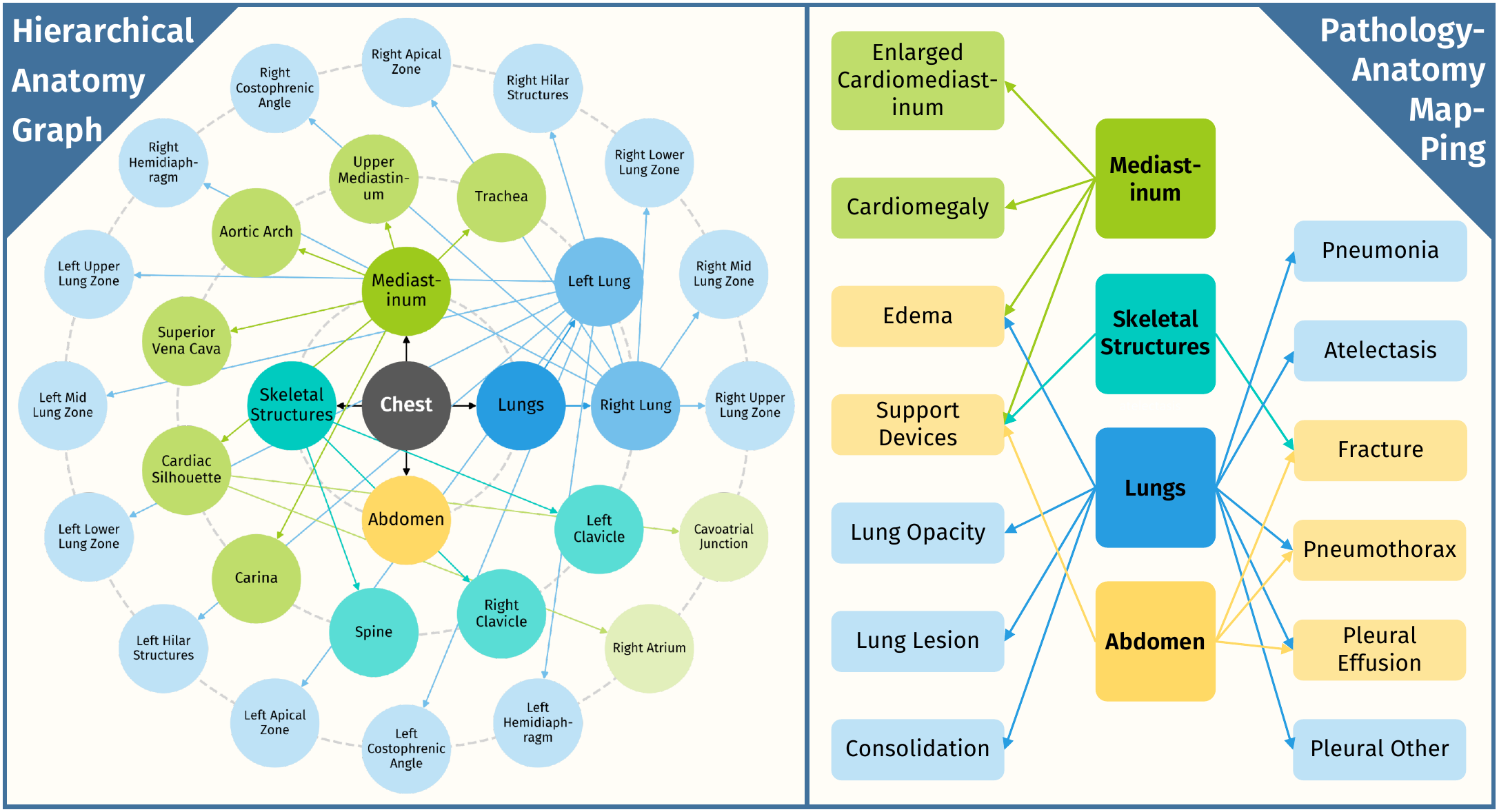}
    \vspace{-5mm}
    \caption{The proposed hierarchical anatomy graph along with the mapping between pathologies and anatomies.}
    \Description{}
    \label{fig:ana-tree}
\end{figure}

\noindent \textbf{Hierarchical Anatomy Graph}.
Conventional graph-based medical report generation often treats anatomical regions at different spatial scales—such as “lungs” and “left lung”—as independent nodes, overlooking their natural inclusion relationships. 
This flat representation leads to semantic ambiguity during alignment, where a phrase like “left pneumonia” may be erroneously assigned to both “lungs” and “left lung.” 
To resolve this, we propose a hierarchical anatomy graph that explicitly encodes spatial containment via directed edges between anatomical regions. 
\nzk{As illustrated in Figure~\ref{fig:ana-tree}, we design a four-level hierarchy tailored to chest X-ray interpretation. }
Inspired by prior anatomical frameworks~\cite{jain2021radgraph, tanida2023rgrg}, the graph begins with four coarse-grained (secondary) regions, which are recursively subdivided into finer-grained (third and fourth-level) regions, resulting in 32 nodes overall—25 of which serve as leaf nodes used for local attention prompting. Given that radiology reports are typically pathology-centered, directly aligning sentences to anatomical entities is non-trivial. To bridge this gap, we curate a pathology-to-anatomy mapping that links reported abnormalities to corresponding leaf-level anatomical regions, using secondary-level anatomies as anchors for intermediate reasoning. 
The \nzk{resulting} graph structure serves as a foundation for anatomy-specific visual queries in subsequent stages, during both training and test phases.

\noindent \textbf{Graph Query Restoration}. 
Given a predefined hierarchical graph $G=(V,E)$, where $V=\{v_i\}_{i=1}^{h}$ denotes the set of anatomical nodes and $h$ is the total number of nodes, and $E=\{(v_i, v_j)|v_i,v_j \in V;v_j \in \mathcal{C}(v_i)\}$ encodes directed edges from each parent node $v_i$ to its children $v_i\in \mathcal{C}(v_i)$, we aim to derive semantically grounded query tokens for both leaf and parent nodes. We begin by encoding each leaf node—identified by an out-degree of zero—using a BERT-based text encoder, yielding the leaf-level query set $L_q=\{q_{v_i}|v_i \in V; {\rm deg}^+(v_i)=0\}$. To obtain the query token for any parent node $p$, we average the representations of its children nodes:
\begin{equation}
    q_p = \frac{1}{|\mathcal{C}(p)|}\sum_{v_i\in \mathcal{C}(p)}q_{v_i},
\end{equation}
where $\mathcal{C}(p)$ denotes the children of node $p$. After recursively restoring all parent node embeddings from their corresponding child nodes, we concatenate the full set of queries—both leaf and reconstructed parent queries—and input them into the visual encoder as anatomical prompts to guide localized attention (Figure~\ref{fig:model} (a)). To preserve anatomical specificity during this process, we apply token-wise masking across prompt tokens during encoding. This ensures that each anatomy query remains isolated from others, preventing premature information fusion and maintaining purity for subsequent anatomical reconstruction.

\nzk{This approach preserves the inherent top-down inclusion relationships between parent and child anatomical structures through average pooling-based restoration, while ensuring the independence of each anatomical query during visual encoding via a simple masking strategy that prevents cross-token information leakage.}

\subsection{Hierarchical Anatomy Reconstruction}
\label{sec:method:2:afr}
\nzk{To capture fine-grained anatomical details and their hierarchical relationships}, we propose an annotation-free hierarchical region reconstruction mechanism that leverages the fused anatomical prompt queries as spatial priors.
Inspired by recent advances in masked image modeling~\cite{he2022mae, zhou2023mrm, huang2024maco}, we begin with global masked image reconstruction. Specifically, given a raw image, we randomly mask a portion of its visual tokens and input them into the visual encoder along with the anatomical prompt queries. Through the query fusion process, we obtain a joint representation \nzk{that includes} the prompt queries, a class token, and the remaining image tokens. These are then passed into a visual decoder to reconstruct the pixel values of the masked regions. The reconstruction loss is computed as the mean squared error (MSE) between the predicted and ground-truth pixel values:
\begin{equation}
    \mathcal{L}_{\rm ir} = \mathbb{E}_{I^{\rm mask}\sim batch}\mathbb{E}_{y_i \in I^{\rm mask}} \| y_i-\hat{y}_i \|_2^2,
    \label{eq:loss-ir}
\end{equation}
where $I^{\rm mask}$ represents the masked input image within a mini-batch, $y_i$ denotes the reconstructed pixel value, and $\hat{y}_i$ is the corresponding ground-truth pixel. The visual decoder follows the same architecture as the encoder, differing only in its output dimension to accommodate pixel reconstruction.

\nzk{While global reconstruction provides coarse supervision, reconstructing local anatomical regions remains a challenge in self-supervised settings due to the absence of expert annotations.} 
To address this, we draw inspiration from synthetic augmentation strategies in medical imaging~\cite{wang2025synthetic} and treat the reconstructed pixel values from unmasked regions as pseudo-labels for their corresponding masked anatomical parts. 
Following the anatomical prompt fusion, each token in the prompt query captures distinct image features aligned with specific anatomical entities. 
\nzk{As a result, even without direct supervision, the fused representations can effectively guide region-specific reconstructions.} 
Leveraging this property, we reuse the visual decoder from the global reconstruction stage to perform localized reconstructions for individual anatomical queries.

This hierarchical anatomy reconstruction process mirrors the earlier graph query restoration, as shown in Figure~\ref{fig:model}(b). Specifically, for a given parent region, we concatenate the masked representations of its children queries and decode them to generate the reconstructed pixel patch. Formally, given the set of fused prompt queries with masked image tokens $\boldsymbol{q}^{\rm mask}=\{q_{v_i}^{\rm mask}|v_i\in V\}$, the reconstructed pixel patch for a parent region $p$ is computed as:
\begin{equation}
    \boldsymbol{y}_p^{\rm mask} = {\rm VD}({\rm concate}(\{q_{v_i}^{\rm mask} | v_i \in \mathcal{C}(p)\})),
\end{equation}
where $\boldsymbol{y}_p^{\rm mask}\in \mathbb{R}^{\omega\times\omega\times d}$ denotes the reconstructed pixel values, $\omega$ is the patch size, $d$ is the channel dimension, and $\rm VD(\cdot)$ is the visual decoder. To generate the pseudo-labels, we decode the unmasked parent query $q_p$ directly: $\boldsymbol{\hat{y}}_p={\rm VD}(q_p)$, which serves as the reconstruction target for the masked input. The overall objective for hierarchical anatomy reconstruction is then formulated as: 
\begin{equation}
    \mathcal{L}_{\rm ar} = \mathbb{E}_{\boldsymbol{q}\sim batch}\mathbb{E}_{q_{p} \in \boldsymbol{q}} \|\boldsymbol{y}_p^{\rm mask}-\boldsymbol{\hat{y}}_p\|_2^2.
    \label{eq:loss-ar}
\end{equation}

\subsection{Anatomical Consistency Alignment}
\label{sec:method:3:acl}

While the proposed hierarchical reconstruction captures top-down anatomical inclusion, it can suffer from trivial solutions. In particular, if the leaf-level prompt tokens become overly similar, their parent representations—obtained via average pooling—will also converge, leading to indistinguishable query representations and ultimately trivial reconstructions where all regions yield near-identical pixel outputs. 
\nzk{This issue stems from weak supervision in pseudo-labels and can undermine the model’s ability to capture the distinct semantics of different anatomical structures.
To mitigate this, we introduce a straightforward yet effective contrastive learning mechanism aimed at fostering diversity among leaf tokens. Specifically, we enforce distinctiveness among leaf-level tokens while ensuring anatomical consistency across samples.}

\nzk{As shown in Figure~\ref{fig:model} (c), we extract the $M$ leaf tokens from each of the $N$ samples in a mini-batch based on the fused prompt queries from unmasked images. 
We then compute a similarity matrix $S^{L} \in \mathbb{R}^{N \times N \times M \times M}$, where each element $s_{ijab}$ represents the cosine similarity between the $a$-th leaf token of sample $i$ and the $b$-th leaf token of sample $j$. 
Building on prior studies on anatomical consistency~\cite{haghighi2022dira, liu2025afire}, we assume that identical anatomical regions across different patients should have similar representations, while distinct regions should remain distinguishable. 
Therefore, we use an identity matrix $\boldsymbol{I}_M$ as the supervision target, aligning only identical anatomical structures across images.
However, this anatomical consistency assumption can be disrupted by non-overlapping abnormalities, which introduce sample-specific deviations and may distort the similarity structure among anatomical tokens. 
To address this, we incorporate textual similarity into the alignment objective, using it as a soft label to refine the target structure. Specifically, we define a textual similarity matrix $S^T \in \mathbb{R}^{N \times N}$, where each entry $s_{ij}^T$ denotes the cosine similarity between the global textual embeddings of reports $T_i$ and $T_j$. 
For the $(i,j)$ block in the anatomical similarity matrix $S^L$, which compares leaf-level tokens between samples $i$ and $j$, we use a weighted soft target matrix: $\boldsymbol{Y_{ij}}=\lambda s_{ij}^T + (1-\lambda)\boldsymbol{I}_M$, which encourages alignment between semantically similar anatomical tokens while allowing flexibility in the presence of abnormalities.}

To ensure that the textual embeddings accurately capture abnor\-mality-related semantics, we adopt a global image-text contrastive loss following \cite{huang2024maco}. Specifically, we align masked images $I_i^{\rm mask}$ with their corresponding reports $T_i$ using the following loss:
\begin{equation}
    \mathcal{L}_{\mathrm{cl}}=\frac{1}{N}\sum_{i=1}^N\log{(\frac{{\rm exp}(\phi(g_i^{I^{\rm mask}},g_i^{T})/\tau)}{\sum_j{\rm exp}(\phi(g_i^{I^{\rm mask}},g_j^{T})/\tau)})},
    \label{eq:loss-cl}
\end{equation}
where $N$ is the batch size, $\phi(\cdot)$ denotes the cosine similarity, and $\tau$ is a learnable temperature parameter~\cite{huang2024maco}. $g_i^{I^{\rm mask}}$ and $g_i^T$ represent the global embeddings of the $i$-th masked image and report, respectively. Building on this, the final anatomical consistency alignment loss is formulated as:
\begin{equation}
    \mathcal{L}_{\rm ac} = \frac{1}{MN^2}\sum_{i=1}^N\sum_{j=1}^N\sum_{a=1}^M(\lambda s_{ij}^T+1-\lambda)\log{(\frac{{\rm exp}(s_{ijaa}^L/\tau)}{\sum_b^M{\rm exp}(s_{ijab}^L/\tau)})},
    \label{eq:loss-ac}
\end{equation}
where $\lambda$ \nzk{balances the influence of the textual similarity-based soft label.} 
This loss promotes intra-class compactness and inter-class separability for anatomical tokens while dynamically adapting to abnormalities across samples.


\subsection{Anatomy-Guided Report Generation}
\label{sec:method:4:agrg}


\nzk{Refining visual representation learning through the invariant top-down inclusion of anatomical structures enables the language model to generate more fine-grained descriptions from disentangled anatomical representations.}

\noindent \textbf{Anatomy Inference}. 
\nzk{We begin by specifying anatomical entities for each image to guide the language model’s focus.}
During training, pathology entities are extracted from ground truth reports using a pretrained knowledge extractor (e.g., CheXbert~\cite{smit2020chexbert}). For testing, we fine-tune a simple linear head on top of the pretrained visual encoder using a linear probing protocol~\cite{zhou2022refers, li2024mlip} to predict pathology entities from images based on silver entities extracted from the ground-truths. In both stages, we map extracted pathology entities to anatomical entities in the hierarchy (Figure~\ref{fig:ana-tree}), linking them to secondary nodes in the anatomical graph and feeding the leaf nodes into the model for report generation.
\nzk{Notably, the provided pathology entities may indicate either normal or abnormal findings.} 
Unlike previous works~\cite{wu2023medklip, tanwani2022repsnet}, which explicitly classify abnormalities before report generation, we use pathologies solely to define the region of focus, avoiding the need to generate reports based on explicit diagnostic results.


\noindent \textbf{Anatomy-Guided Report Generation}. 
To facilitate parallel generation, we construct the complete anatomical prompt query along with the predicted pathology entities as the prefix. We then set the attention masks based on the presence of each anatomy and pathology, allowing the model to focus on the relevant parts of the input. Consequently, the loss function for report generation is formulated as follows:
\begin{equation}
    \mathcal{L}_{\rm lm} = -\mathbb{E}_{(I,T)\sim batch}\sum^{|T|}_{i=1}\log{p(t_i|I,P_I,t_{<i})},
    \label{eq:loss-lm}
\end{equation}
where $t_i$ denotes the language token at position $i$, $t_{<i}$ represents the tokens generated before $t_i$, $|T|$ is the length of the report tokens, and $P_I$ denotes the combined anatomy-pathology prefix for image $I$. The final loss function for training SS-ACL combines the prior losses, weighted by the respective coefficients:
\begin{equation}
    \mathcal{L}=\alpha_1(\mathcal{L}_{\rm ir} + \mathcal{L}_{\rm ar}) + \alpha_2(\mathcal{L}_{\rm ac} + \mathcal{L}_{\rm cl}) + \alpha_3\mathcal{L}_{\rm lm}.
    \label{eq:loss-final}
\end{equation}

\section{Experiment}
\label{sec:eval}

\setlength{\tabcolsep}{2pt}
\begin{table*}[]
\centering
\caption{Comparison of fine-tuned report generation with SOTAs on the MIMIC-CXR test set, evaluated using NLG, CE, graph-based and vector metrics. $\downarrow$ indicates lower is better. $*$ denotes results obtained by rerunning the original inference code and computing scores using the code from~\cite{yu2023evaluating}. $\dagger$ indicates results quoted from~\cite{jeong2024xrem}. $\uparrow \Delta$ indicates the average improvement ratio of our method relative to previous works. Rad-F1 and CheXbert are shorts for the radgraph F1 score and the CheXbert vector similarity. The best and second-best results within the same test set as ours are highlighted in {\color[HTML]{C00000} \textbf{red}} and {\color[HTML]{4472C4} \textbf{blue}}, respectively.}
\vspace{-3mm}
\label{tab:mrg-mimic-sota}
\resizebox{\textwidth}{!}{%
\begin{tabular}{@{}lccccccccccccccr@{}}
\toprule
 &
   &
   &
  \multicolumn{3}{c}{NLG Metrics} &
   &
  \multicolumn{3}{c}{CE Metrics} &
   &
  \multicolumn{3}{c}{Graph \& Vector   Metrics} &
   &
   \\ \cmidrule(lr){4-6} \cmidrule(lr){8-10} \cmidrule(lr){12-14}
\multirow{-2}{*}{Models} &
  \multirow{-2}{*}{\begin{tabular}[c]{@{}c@{}}Param. Count\\ Vision / LM\end{tabular}} &
  \multirow{-2}{*}{Year} &
  BLEU-4 &
  ROUGE &
  METEOR &
  \multirow{-2}{*}{$\uparrow \Delta$} &
  Precision &
  Recall &
  F1 &
  \multirow{-2}{*}{$\uparrow \Delta$} &
  Rad-F1 &
  RadCliQ ($\downarrow$) &
  CheXbert &
  \multirow{-2}{*}{$\uparrow \Delta$} &
  \multirow{-2}{*}{Test Set Size} \\ \midrule

CXR-RePaiR~\cite{endo2021cxr-repair} &
  86M / 63M &
  2021 &
  - &
  - &
  - &
  - &
  - &
  - &
  - &
  - &
  0.090 &
  4.12 &
  0.379 &
  - &
  2192 samples \\
X-REM~\cite{jeong2024xrem} &
  86M / 124M &
  2023 &
  - &
  - &
  - &
  - &
  - &
  - &
   &
  - &
  0.172 &
  3.84 &
  0.351 &
  - &
  2192 samples \\
Med-PaLM M~\cite{tu2024medpalmm} &
  22B / 66B &
  2023 &
  0.133 &
  - &
  0.296 &
  - &
  - &
  - &
  0.398 &
  - &
  0.267 &
  - &
  - &
  - &
  2347 reports \\
COMG~\cite{gu2024comg} &
  46M / 41M &
  2024 &
  0.104 &
  0.279 &
  0.137 &
  - &
  0.424 &
  0.291 &
  0.345 &
  - &
  - &
  - &
  - &
  - &
  2065 reports \\
ControlDiff~\cite{tian2024controldiff} &
  86M / - &
  2024 &
  0.132 &
  0.299 &
  0.186 &
  - &
  0.477 &
  0.484 &
  0.480 &
  - &
  - &
  - &
  - &
  - &
  52,000 images \\
MAIRA-1~\cite{hyland2023maira} &
  87M / 7B &
  2024 &
  0.142 &
  0.333 &
  0.289 &
  - &
  - &
  - &
  0.386 &
  - &
  0.243 &
  3.10 &
  0.440 &
  - &
  2461 samples \\ \midrule
WCL$^{\dagger}$~\cite{yan2021wcl} &
  46M / 41M &
  2021 &
  0.067 &
  0.241 &
  0.100 &
  76\% &
  0.385 &
  0.274 &
  0.294 &
  66\% &
  {\color[HTML]{4472C4} \textbf{0.143}}$^*$ &
  {\color[HTML]{4472C4} \textbf{3.99}}$^*$ &
  {\color[HTML]{4472C4} \textbf{0.309}}$^*$ &
  {\color[HTML]{C00000} \textbf{32\%}} &
   \\
R2GenCMN~\cite{chen2021r2gencmn} &
  46M / 41M &
  2022 &
  0.106 &
  0.278 &
  0.142 &
  26\% &
  0.334 &
  0.275 &
  0.278 &
  77\% &
  0.117$^*$ &
  4.18$^*$ &
  0.210$^*$ &
  67\% &
   \\
METransformer~\cite{wang2023metransformer} &
  86M / 66M &
  2023 &
  0.124 &
  0.291 &
  0.152 &
  15\% &
  0.364 &
  0.309 &
  0.334 &
  56\% &
  - &
  - &
  - &
  - &
   \\
RGRG~\cite{tanida2023rgrg} &
  49M / 355M &
  2023 &
  0.121$^*$ &
  0.283$^*$ &
  0.160$^*$ &
  15\% &
  0.384$^*$ &
  0.367$^*$ &
  0.338$^*$ &
  45\% &
  0.131$^*$ &
  4.11$^*$ &
  0.223$^*$ &
  56\% &
   \\
R2GenGPT~\cite{wang2023r2gengpt} &
  87M / 7B &
  2023 &
  {\color[HTML]{4472C4} \textbf{0.132}} &
  {\color[HTML]{4472C4} \textbf{0.300}} &
  0.158 &
  {\color[HTML]{C00000} \textbf{10\%}} &
  0.310 &
  0.345 &
  0.321 &
  62\% &
  0.123$^*$ &
  4.09$^*$ &
  0.224$^*$ &
  59\% &
   \\
Med-LLM~\cite{liu2024medllm} &
  307M / 7B &
  2024 &
  0.128 &
  0.289 &
  0.161 &
  12\% &
  0.412 &
  0.373 &
  0.395 &
  33\% &
  - &
  - &
  - &
  - &
   \\
DCG~\cite{liang2024dcg} &
  86M / 82M &
  2024 &
  0.126 &
  0.295 &
  {\color[HTML]{4472C4} \textbf{0.162}} &
  11\% &
  {\color[HTML]{4472C4} \textbf{0.441}} &
  {\color[HTML]{4472C4} \textbf{0.414}} &
  {\color[HTML]{4472C4} \textbf{0.404}} &
  {\color[HTML]{C00000} \textbf{25\%}} &
  - &
  - &
  - &
  - &
   \\
\textbf{SS-ACL} &
  86M / 356M &
  \textbf{ours} &
  {\color[HTML]{C00000} \textbf{0.144}} &
  {\color[HTML]{C00000} \textbf{0.319}} &
  {\color[HTML]{C00000} \textbf{0.180}} &
  - &
  {\color[HTML]{C00000} \textbf{0.603}} &
  {\color[HTML]{C00000} \textbf{0.468}} &
  {\color[HTML]{C00000} \textbf{0.505}} &
  - &
  {\color[HTML]{C00000} \textbf{0.225}} &
  {\color[HTML]{C00000} \textbf{3.23}} &
  {\color[HTML]{C00000} \textbf{0.489}} &
  - &
  \multirow{-9}{*}{3858 samples}
 \\ \bottomrule
\end{tabular}%
}
\end{table*}

\subsection{Tasks and Datasets}

\noindent \textbf{Tasks.} To evaluate the synergistic capabilities of SS-ACL across both vision and language tasks, we perform several evaluations: \textbf{Report Generation} for language performance, zero-shot and fine-tuning \textbf{Image Classification}, and fine-tuning \textbf{Segmentation} for isolated vision tasks. 
\nzk{To further examine vision-language synergy, we evaluate zero-shot \textbf{Phrase Grounding} and \textbf{Image-to-Text Retrieval}.}

\noindent \textbf{Datasets.}
\nzk{We conduct experiments on the MIMIC-CXR dataset~\cite{johnson2019mimic}, following prior work~\cite{chen2020r2gen, wang2023r2gengpt}, using 222,758/1,808/3,858 images for training, validation, and testing, respectively.}
For downstream evaluations, we use the NIH ChestX-ray~\cite{wang2017nih} and RSNA Pneumonia~\cite{shih2019rsna} datasets for image classification, SIIM-ACR Pneumothorax~\cite{siim-acr-pneumothorax-segmentation} for segmentation, CheXpert 5x200~\cite{huang2021gloria, zhou2023mrm} for retrieval, and MS-CXR~\cite{boecking2022biovil} for phrase grounding. 
\nzk{Additional dataset details are provided in the Supplementary Material.}

\subsection{Evaluation Metrics}

\noindent \textbf{Natural Language Generation (NLG) Metrics.}
To evaluate lexical accuracy, we report BLEU~\cite{Papineni2002BLEU}, METEOR~\cite{Denkowski2014Meteor}, and ROUGE~\cite{Lin2004Rouge:} scores, as in~\cite{wang2023r2gengpt, liu2024medllm}, \nzk{using human-written reports as references.}

\noindent \textbf{Clinical Efficacy (CE) Metrics.}
For assessing the pathological accuracy of generated reports, we report macro-averaged scores for Precision, Recall, and F1-score, following works~\cite{chen2020r2gen, wang2023r2gengpt, wang2023metransformer}. 
Generated reports are labeled using CheXbert~\cite{smit2020chexbert} across 14 categories of thoracic diseases, with the ``uncertain" label treated as negative.

\noindent \textbf{Graph and Vector Metrics.}
Following~\cite{jeong2024xrem, tu2024medpalmm, hyland2023maira}, we compute graph-based metrics (Radgraph-F1 and RadCliQ v0) and CheXbert vector similarity between generated reports and references using the code from~\cite{yu2023evaluating}. These metrics capture clinically relevant aspects, focusing on content accuracy rather than superficial phrasing variations~\cite{hyland2023maira}.

\noindent \textbf{Metrics for Downstream Tasks.}
For classification, we report the area under the receiver operating characteristic curve (AUC) score. For retrieval, we measure semantic correlation between images and reports using the precision score, denoted as Prec@$K$, where $K$ is the number of retrieved reports corresponding to a given image. For segmentation, we report the Dice score.

\subsection{Implementation Details}
\label{sec:expsettings}

\noindent \textbf{Pre-training.}
\nzk{Following prior works}~\cite{huang2024maco, tanida2023rgrg}, we use the standard ViT-B with a patch size of 16 as the visual encoder and fine-tune GPT-2 Medium~\cite{radford2019gpt2} on PubMed abstracts~\cite{papanikolaou2020dare} as the language decoder. For the text encoder, we adopt a 12-layer BERT~\cite{kenton2019bert} with an inner projection dimension of 384. The hyperparameter $\lambda$ in Equation~\ref{eq:loss-ac} is set to 0.1, and the coefficients $[\alpha_1,\alpha_2,\alpha_3]$ for each loss in Equation~\ref{eq:loss-final} are set to $[1.0, 0.15, 1.0]$, respectively. 
\nzk{Our implementation builds upon the publicly available MaCo codebase\footnote{\url{https://github.com/SZUHvern/MaCo}} using PyTorch, and all joint pre-training experiments are conducted on four NVIDIA A6000 GPUs with a total batch size of $128$ for $40$ epochs.} The initial learning rate is set to $2.5 \times 10^{-4}$. Additional experimental details are provided in the Supplementary Material.

\noindent \textbf{Downstream Fine-tuning.}
\nzk{We evaluate SS-ACL across various downstream tasks and datasets. For classification and segmentation, we follow the MRM framework~\cite{zhou2023mrm}; retrieval tasks are implemented using GLoRIA~\cite{huang2021gloria}; and phrase grounding is conducted with the MaCo pipeline~\cite{huang2024maco}. Further implementation details are consistent with the settings provided in the respective original works.}


\subsection{Comparison with State-of-The-Arts}
\label{sec:eval_r2g}

\noindent \textbf{Report Generation.}
\nzk{Recent advances in radiology report generation have explored diverse strategies for integrating medical knowledge~\cite{wang2022msat, tanida2023rgrg, wang2023metransformer, gu2024comg}, employing retrieval-augmented methods~\cite{endo2021cxr-repair, jeong2024xrem}, and leveraging the transfer capabilities of large language models~\cite{wang2023r2gengpt, liu2024bllm, tu2024medpalmm, hyland2023maira}. }
To evaluate generation performance, we compare our SS-ACL with state-of-the-art models from the past five years using three types of metrics. As shown in Table~\ref{tab:mrg-mimic-sota}, our model outperforms current SOTA by an average of \textbf{10}\%, \textbf{25}\%, and \textbf{32}\% in NLG, CE, and Graph metrics, respectively.

\begin{table}[]
\centering
\caption{Comparison of AUC scores (\%) for image classification on NIH and RSNA with varying ratios of annotated samples. $\dagger$ denotes models pretrained using our training split.}
\label{tab:cls}
\resizebox{\linewidth}{!}{%
\begin{tabular}{@{}lcccccccc@{}}
\toprule
 &
  \multicolumn{4}{c}{NIH (AUC)} &
  \multicolumn{4}{c}{RSNA (AUC)} \\ \cmidrule(l){2-9} 
\multirow{-2}{*}{Models} &
  Zero-shot &
  1\% &
  10\% &
  100\% &
  Zero-shot &
  1\% &
  10\% &
  100\% \\ \midrule
GLoRIA~\cite{huang2021gloria} &
  61.0 &
  67.1 &
  76.6 &
  81.3 &
  80.4 &
  86.1 &
  88.0 &
  88.6 \\
ConVIRT~\cite{zhang2022convirt} &
  66.1 &
  66.2 &
  76.6 &
  81.3 &
  71.5 &
  77.4 &
  80.1 &
  81.3 \\
BioViL~\cite{boecking2022biovil} &
  69.1 &
  69.5 &
  75.3 &
  82.5 &
  82.8 &
  88.1 &
  88.4 &
  89.1 \\
REFERS~\cite{zhou2022refers} &
  {\color[HTML]{4472C4} \textbf{-}} &
  76.7 &
  80.9 &
  84.7 &
  {\color[HTML]{4472C4} \textbf{-}} &
  89.4 &
  91.6 &
  92.7 \\
MedKLIP~\cite{wu2023medklip} &
  {\color[HTML]{C00000} \textbf{76.8}} &
  77.2 &
  78.9 &
  83.2 &
  86.9 &
  87.3 &
  88.0 &
  89.3 \\
MRM~\cite{zhou2023mrm}$^{\dagger}$ &
  - &
  77.3 &
  {\color[HTML]{C00000} \textbf{83.5}} &
  {\color[HTML]{4472C4} \textbf{85.5}} &
  - &
  {\color[HTML]{C00000} \textbf{91.0}} &
  {\color[HTML]{4472C4} \textbf{91.8}} &
  91.9 \\
MaCo~\cite{huang2024maco}$^{\dagger}$ &
  {\color[HTML]{4472C4} \textbf{74.4}} &
  {\color[HTML]{C00000} \textbf{78.1}} &
  {\color[HTML]{4472C4} \textbf{83.4}} &
  85.4 &
  {\color[HTML]{4472C4} \textbf{87.5}} &
  89.7 &
  {\color[HTML]{4472C4} \textbf{91.8}} &
  {\color[HTML]{4472C4} \textbf{92.0}} \\
\textbf{SS-ACL} &
  71.9 &
  {\color[HTML]{C00000} \textbf{78.1}} &
  {\color[HTML]{4472C4} \textbf{83.4}} &
  {\color[HTML]{C00000} \textbf{85.6}} &
  {\color[HTML]{C00000} \textbf{88.7}} &
  {\color[HTML]{4472C4} \textbf{90.8}} &
  {\color[HTML]{C00000} \textbf{91.9}} &
  {\color[HTML]{C00000} \textbf{92.9}} \\ \bottomrule
\end{tabular}%
}
\end{table}

\noindent \textbf{Image Classification and Segmentation.}
Table~\ref{tab:cls} presents the image classification results for various pretraining models, evaluating the generalizability of visual representations in both zero-shot and fine-tuning scenarios with different training sample ratios. SS-ACL performs comparably to state-of-the-art models across most scenarios, even outperforming the current strongest foundation model, MaCo~\cite{huang2024maco}, on the RSNA dataset with limited samples. 
\nzk{These results underscore the effectiveness of our anatomical consistency learning strategy in producing robust and transferable visual features.}

\nzk{Table~\ref{tab:pg-retr-seg} reports segmentation results after fine-tuning various pretraining models, providing further insights into the capacity of pre-trained representations to localize pathological regions}
SS-ACL surpasses SOTA foundation models, demonstrating its ability to provide grounded information for downstream tasks.

\noindent \textbf{Phrase Grounding and Retrieval.}
\nzk{To assess its multi-modal synergy effectiveness—specifically, its capacity to ground visual evidence within generated reports—we conduct zero-shot phrase grounding and image-to-text retrieval following established protocols~\cite{huang2021gloria, huang2024maco}. 
As shown in Table~\ref{tab:pg-retr-seg}, SS-ACL outperforms prior SOTA models.}

\begin{figure}[t]
    \centering
    \includegraphics[width=\linewidth]{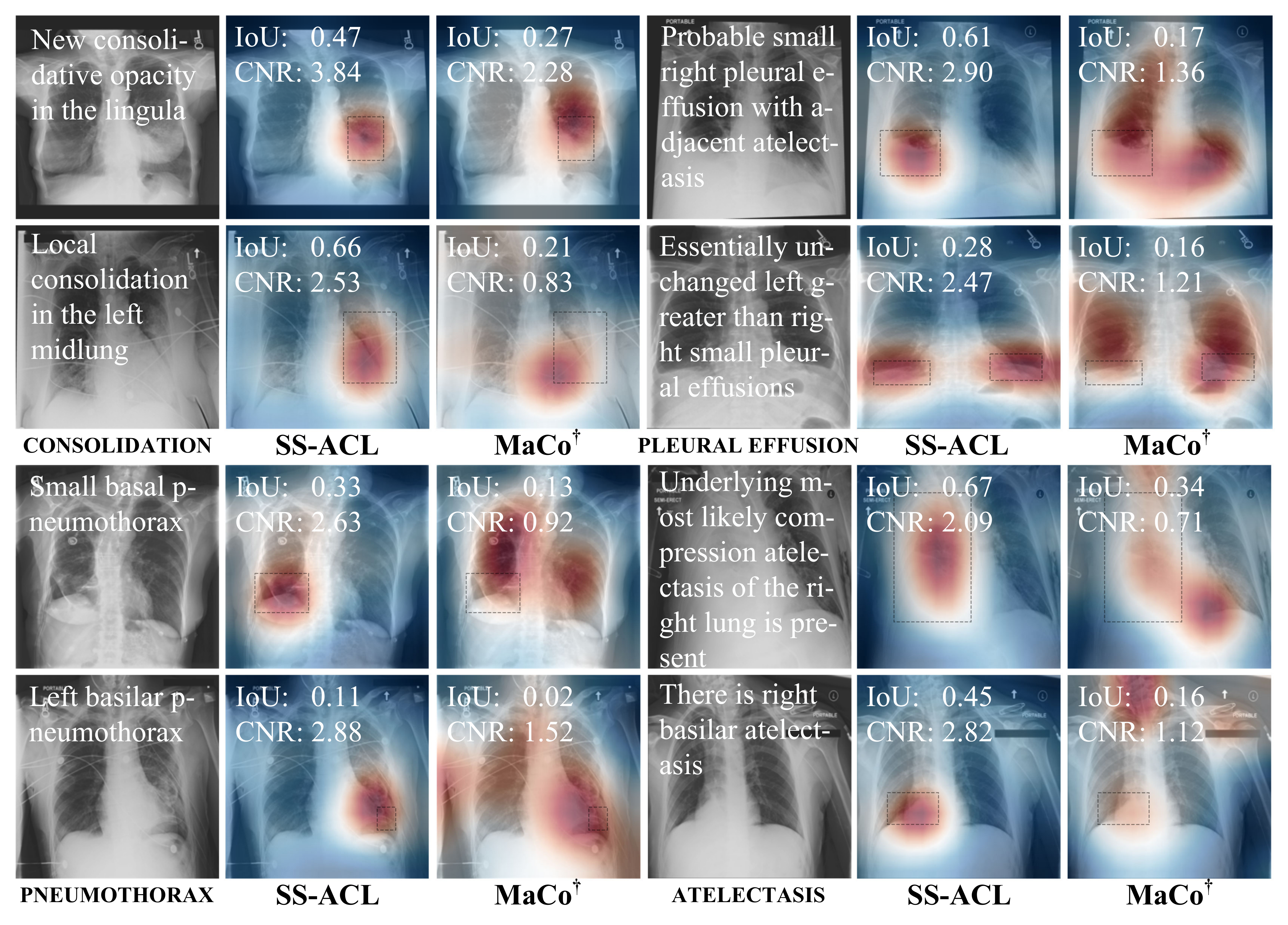}
    \caption{Qualitative phrase-grounding results. Given description phrases, we visualize the association of vision and language using the DINO attention map on the MS-CXR dataset. Ground-truth annotations are represented with dashed boxes. $\dagger$ denotes models pretrained using our training split.}
    \Description{}
    \label{fig:case-study-pg}
\end{figure}

\nzk{For phrase grounding, we construct a weighted patch-wise attention map between the class token and image patches using the DINO attention mechanism~\cite{caron2021dino}, and compute the cross-modal correlation between global sentence and image embeddings. }
In Figure~\ref{fig:case-study-pg}, we visualize the phrase grounding results for various pathologies. Compared to the SOTA model MaCo, SS-ACL produces more accurate and focused attention maps, particularly for small and multi-located pathologies (e.g., pneumothorax in the last row and pleural effusion in the second row). 
\nzk{These improvements highlight how our hierarchical anatomical reconstruction fosters fine-grained structural representations that better align visual regions with textual descriptions.}

For image-to-text retrieval, SS-ACL also shows stronger performance under the zero-shot protocol~\cite{huang2021gloria}. 
As summarized in Table~\ref{tab:pg-retr-seg}, \nzk{top-$K$ candidate reports are ranked based on their similarity to a query image, and precision is computed by comparing predicted pathological labels from the retrieved reports to ground truth annotations derived from CheXbert~\cite{smit2020chexbert}.}
The results demonstrate synergistic improvements in both visual diagnosis and report generation. 
\nzk{These findings collectively suggest that \textbf{SS-ACL presents a promising direction for extending conventional unimodal medical foundation models into robust multi-modal frameworks}, enabling accurate and interpretable clinical decision-making across tasks.}


\begin{table}[t]
\centering
\caption{Comparison for zero-shot phrase grounding on the MS-CXR dataset, zero-shot image-to-text retrieval on the CheXpert 5x200 dataset, and fine-tuning image segmentation on the SIIM dataset. $\dagger$ denotes models pretrained using our training split.}
\label{tab:pg-retr-seg}
\resizebox{\linewidth}{!}{%
\begin{tabular}{@{}lccccc@{}}
\toprule
                         & \multicolumn{2}{c}{Phrase Grounding}                                          & \multicolumn{2}{c}{Image-to-Text Retrieval} & Segmentation \\ \cmidrule(l){2-6} 
\multirow{-2}{*}{Models} & CNR                                   & mIoU                                  & Prec@5               & Prec@10              & Dice         \\ \midrule
GLoRIA~\cite{huang2021gloria}                   & 0.93                                  & {\color[HTML]{C00000} \textbf{0.246}} & 32.6                 & 33.4                 & 63.4         \\
ConVIRT~\cite{zhang2022convirt}                  & 0.818                                 & {\color[HTML]{4472C4} \textbf{0.238}} & 30.8                 & 28.2                 & 59.9         \\
BioViL~\cite{boecking2022biovil}                   & {\color[HTML]{4472C4} \textbf{1.140}} & 0.215                                 & -                    & -                    & 70.0         \\
MLIP~\cite{li2024mlip}                     & -                                     & -                                     & 39.0                 & 39.4                 & -            \\
MaCo~\cite{huang2024maco}$^{\dagger}$ &
  1.066 &
  0.228 &
  {\color[HTML]{4472C4} \textbf{57.2}} &
  {\color[HTML]{4472C4} \textbf{54.3}} &
  {\color[HTML]{4472C4} \textbf{89.4}} \\
\textbf{SS-ACL} &
  {\color[HTML]{C00000} \textbf{1.153}} &
  0.229 &
  {\color[HTML]{C00000} \textbf{58.3}} &
  {\color[HTML]{C00000} \textbf{55.0}} &
  {\color[HTML]{C00000} \textbf{89.9}} \\ \bottomrule
\end{tabular}%
}
\end{table}

\begin{figure*}[t]
    \centering
    \includegraphics[width=\textwidth]{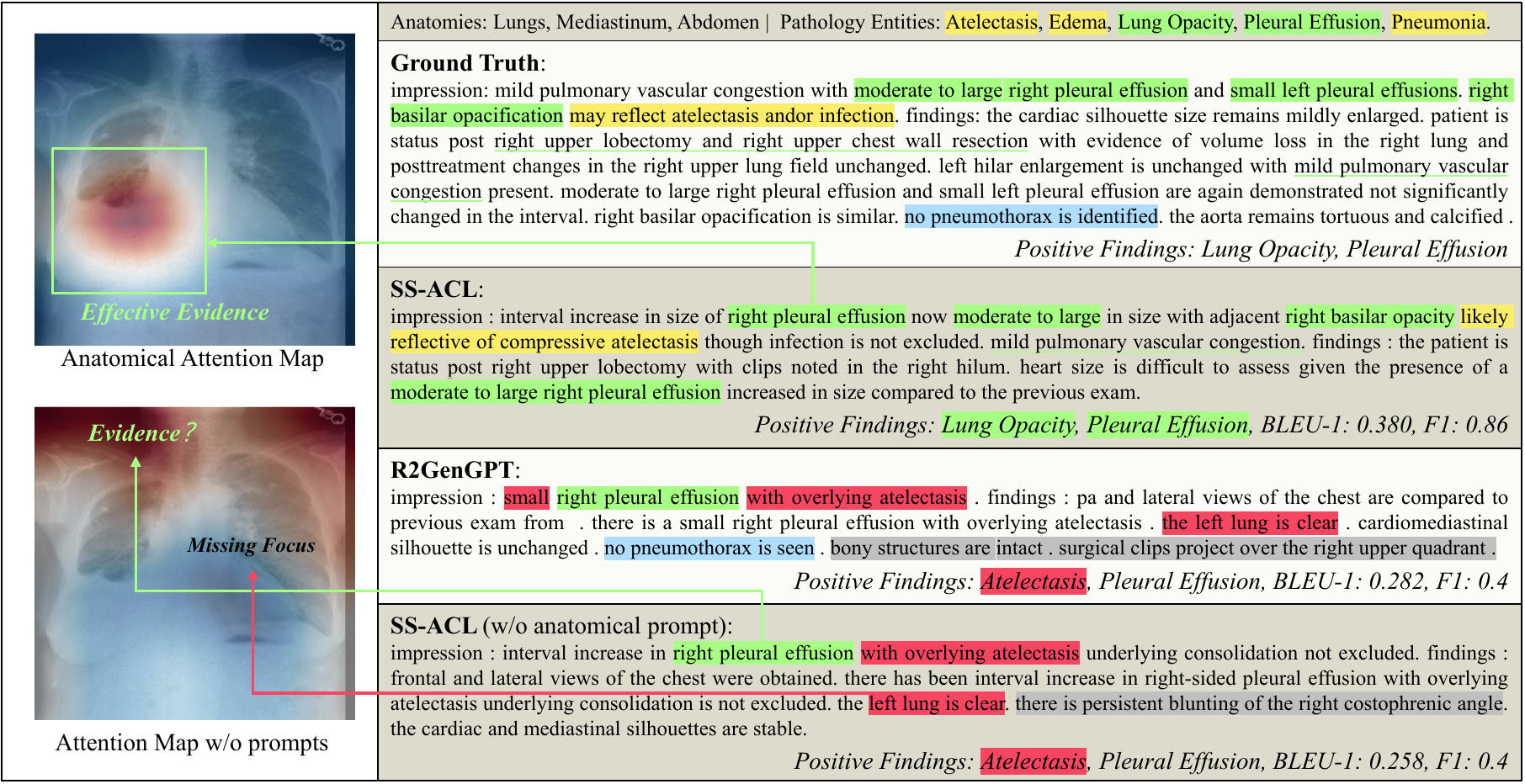}
    \caption{Qualitative report samples generated by SS-ACL with corresponding visual evidence and conventional SOTA models from the MIMIC-CXR test set. The green highlights indicate detected positive abnormalities, green underlines mark undetected positives, yellow highlights denote uncertain pathologies, red marks indicate incorrect estimations, blue marks signify negatives, and gray highlights represent unrelated descriptions.}
    \Description{}
    \label{fig:case-study}
\end{figure*}

\subsection{Ablation Study}
\label{sec:abs}


\begin{table}[]
\centering
\caption{Ablation study of SS-ACL, evaluating the isolated contribution of each sub-module to language and visual downstream tasks.}
\label{tab:ablation}
\resizebox{\linewidth}{!}{%
\begin{tabular}{@{}lcccccc@{}}
\toprule
                                    & \multicolumn{2}{c}{NLG Metrics} & \multicolumn{2}{c}{Graph Metrics} & \multicolumn{2}{c}{Visual Metrics}            \\ \cmidrule(l){2-7} 
\multirow{-2}{*}{Ablation Settings} & BLEU-4         & METEOR         & Rad-F1     & CheXbert    & AUC                                   & CNR   \\ \midrule
Baseline (Encoder-Decoder)          & 0.108          & 0.135          & 0.140           & 0.277              & 0.612                                 & 0.866 \\
+ MAE (Global)                      & 0.118          & 0.152          & 0.160           & 0.405              & 0.691                                 & 0.926 \\
+ CLIP (Global)                     & 0.125          & 0.158          & 0.174           & 0.373              & 0.698                                 & 1.011 \\
+ HAR                               & 0.139          & 0.171          & 0.204           & 0.434              & 0.701                                 & 1.118 \\
\hspace{2em}a) 1 layer Anatomy Graph            & 0.109          & 0.136          & 0.139           & 0.277              & 0.688                                 & 0.936 \\
\hspace{2em}b) 2 layer Anatomy Graph            & 0.123          & 0.156          & 0.172           & 0.392              & 0.698                                 & 1.001 \\
\hspace{2em}c) 3 layer Anatomy Graph            & 0.135          & 0.169          & 0.189           & 0.423              & 0.695                                 & 1.003 \\
\hspace{2em}d) 4 layer Anatomy Graph            & 0.139          & 0.171          & 0.204           & 0.434              & 0.701                                 & 1.118 \\
+ ACA &
  {\color[HTML]{4472C4} \textbf{0.141}} &
  {\color[HTML]{4472C4} \textbf{0.174}} &
  {\color[HTML]{4472C4} \textbf{0.213}} &
  {\color[HTML]{4472C4} \textbf{0.451}} &
  0.718 &
  {\color[HTML]{C00000} \textbf{1.168}} \\
\hspace{2em}- Textual Smoothing                 & 0.140          & 0.173          & 0.192           & 0.402              & 0.693                                 & 1.142 \\
+ Pathology Entity &
  {\color[HTML]{C00000} \textbf{0.144}} &
  {\color[HTML]{C00000} \textbf{0.180}} &
  {\color[HTML]{C00000} \textbf{0.225}} &
  {\color[HTML]{C00000} \textbf{0.489}} &
  {\color[HTML]{4472C4} \textbf{0.719}} &
  {\color[HTML]{4472C4} \textbf{1.153}} \\
\hspace{2em}+ Diagnostic Labels                 & 0.138          & 0.170          & 0.202           & 0.429              & {\color[HTML]{C00000} \textbf{0.723}} & 1.144 \\ \bottomrule
\end{tabular}%
}
\end{table}

To evaluate each component of our framework, we conduct a comprehensive assessment of both language performance for report generation and visual performance in zero-shot scenarios. 
\nzk{As shown in Table~\ref{tab:ablation}, we begin with a standard encoder-decoder baseline composed of ViT-B and GPT-2, matching the architecture of SS-ACL for fair comparison. We incrementally introduce key modules, including global masked image reconstruction (MAE, Eq.~\ref{eq:loss-ir}), global visual-language contrastive learning (CLIP, Eq.~\ref{eq:loss-cl}), hierarchical anatomy reconstruction (HAR, Eq.~\ref{eq:loss-ar}), anatomical consistency alignment (ACA, Eq.~\ref{eq:loss-ac}), and the use of predicted pathology entities as input tokens to the decoder.}
For HAR, we also examine the impact of different scales for the predefined anatomy graph (4 layers), pruning the graph from the leaf to the secondary layer.

The results show that hierarchical reconstruction outperforms global reconstruction, enhancing both language and visual performance. 
\nzk{However, as finer layers are removed, performance deteriorates—underscoring the value of preserving detailed anatomical priors.}
For ACA, we observe improved visual performance, particularly in zero-shot phrase grounding, highlighting the benefits of local anatomical consistency over global contrastive learning. We also test the alignment without the proposed textual smoothing for normal and abnormal samples, confirming that without smoothing, abnormalities disrupt the alignment, supporting our hypothesis. Finally, we compare the use of predicted pathology entities with and without diagnostic labels as input for report generation. As discussed in our methodology, relying too heavily on diagnostic labels degrades the model’s diagnostic capability, reinforcing the importance of maintaining a balanced input.

\subsection{Case Study}
\label{sec:case}

In Figure~\ref{fig:case-study}, we visualize the association between images and reports generated by SS-ACL. Traditional end-to-end report generation frameworks, such as R2GenGPT~\cite{wang2023r2gengpt}, typically evaluate model performance based on language metrics that measure the overlap of words or pathologies between predictions and ground truths. 
\nzk{However, such metrics fail to reveal the root causes of diagnostic errors—particularly, the model's inability to ground language outputs in visual evidence. Our analysis reveals that in the absence of the proposed anatomical query prompt, the model's attention becomes dispersed and fails to focus on diagnostically relevant regions. 
This often results in hallucinated or imprecise descriptions that misrepresent the underlying image content. }

In contrast, when guided by sophisticated anatomical prompts, SS-ACL easily locates the abnormal regions and generates precise descriptions without hallucinations. In the given case, the ground truth describes a positive lung opacity and pleural effusion, with uncertain mentions of atelectasis, edema, and pneumonia. Treating uncertain pathologies as negatives, the model should only describe the lung opacity and pleural effusion. However, the conventional model, lacking anatomical guidance, mistakenly predicts pathologies like atelectasis and irrelevant details such as ``bony structures" and ``surgical clips." On the other hand, SS-ACL, with the correct anatomical guidance, accurately describes the image as per the prompt query. This highlights its ability to control generated content by explicitly specifying anatomical information in the prompt.

Through this case study, we demonstrate that interpretability and model performance in medical diagnosis are inherently linked and can mutually enhance each other, even in the absence of expert annotations. Additional qualitative samples are provided in the Supplementary Materials.

\section{Conclusion}
\label{sec:conclusion}

We introduce Self-Supervised Anatomical Consistency Learning (SS-ACL), an innovative framework for visual-grounded medical report generation. SS-ACL is designed to generate both generalized visual representations and precise pathological descriptions, enhanced by interpretable visual evidence. By leveraging the top-down inclusion relationship inherent in anatomical structures, SS-ACL constructs a hierarchical anatomy graph to guide the visual encoder in reconstructing anatomical focus. The model further improves performance by aligning anatomies across images using a simple yet effective textual prompt, mitigating degradation and providing targeted guidance for report generation. Extensive experiments validate SS-ACL’s exceptional performance across both visual and language tasks. With robust zero-shot capabilities, SS-ACL outperforms current foundation models, particularly in data-scarce settings, and positions itself as a powerful tool for advancing multimodal medical tasks and clinical decision-making.

\section{Acknowledgments} 
This research was supported in part by the National Natural Science Foundation of China under Grant 62171323, 62271155, in part by the Yeqisun Joint Funds of the National Natural Science Foundation of China under Grant U2441252, in part by National Key R\&D Program of China under Grant 2020YFA0711400, in part by Shanghai Municipal Science and Technology Major Project (2021SHZDZX0100), in part by the Changjiang Scholars Program of China, in part by the Computational Biology Program (25JS2840100) of Science and Technology Commission of Shanghai Municipality (STCSM).

\bibliographystyle{ACM-Reference-Format}
\balance
\bibliography{MM25}










\newpage
\appendix
\section*{Supplementary Material}

In this supplementary material, we commence by detailing the experimental settings, encompassing the detailed information of datasets utilized in our experiments (Section~\ref{suppl:sec:dataset}), as presented in Section~\ref{suppl:sec:exp-settings}. Subsequently, Section~\ref{suppl:sec:ablation} introduces additional qualitative ablation analysis aimed at further investigating the inherent discrimination of our pretrained visual representations. Moreover, we provide additional comprehensive qualitative results for zero-shot phrase grounding and pneumonia segmentation, along with an analysis that underscores the synergistic effectiveness of our visual representation and report generation capabilities. Finally, we address the limitations of SS-ACL in Section~\ref{suppl:sec:discussion}. 

We are committed to ensuring the reproducibility of our work and will release the source code and pretraining weights upon publication. Interested researchers will be able to access the code through our designated public repository.

\section{Detailed Experimental Settings}
\label{suppl:sec:exp-settings}

\subsection{Detailed Dataset Information}
\label{suppl:sec:dataset}

In our experiments, several popular benchmarks are adopted as illustrated in Table~\ref{tab:dataset}. For training stage, we use the MIMIC-CXR~\cite{johnson2019mimic} to train our uniform framework. To comprehensively evaluate our model on different downstream tasks, we follow previous work~\cite{huang2021gloria,huang2024maco} to employ additional 4 datasets with distinct sizes and annotations. The detailed information of the splits and annoations is listed in the table. Note that, CheXpert-5x200~\cite{zhang2022convirt,huang2021gloria} is a selected subset of CheXpert~\cite{irvin2019chexpert} for evaluation of zero-shot retrieval, which contains 200 explicitly single-labeled samples for each of the five pathologies mentioned, respectively. Similarly, MS-CXR~\cite{boecking2022biovil} is a manually selected subset of MIMIC-CXR, which contains additional annotations of the pathology bounding boxes and the corresponding pixel-level segmentation map.

\textbf{Training Dataset.} We trained our model on the most popular \textbf{MIMIC-CXR}\footnote{https://physionet.org/content/mimic-cxr/2.0.0/}~\cite{johnson2019mimic} dataset. It contains 377,110 chest X-ray images and 227,835 reports from 64588 patients of the Beth Israel Deaconess Medical Center examined between 2011 and 2016. The popular split is adopted following works~\cite{chen2020r2gen,wang2023metransformer,wang2023r2gengpt} for a fair comparison.



\textbf{CheXpert}~\cite{irvin2019chexpert} is a large-scale medical dataset containing a total of 224,316 chest radiographs from 65,240 patients for multi-label classification tasks. Each radiograph is labeled for the presence of 14 total medical observations. We utilize the selected subset CheXpert-5x200~\cite{zhang2022convirt,huang2021gloria} for evaluation of zero-shot retrieval, which contains 200 explicitly single-labeled samples for each of the five pathologies mentioned, respectively.

\textbf{NIH ChestX-ray}~\cite{wang2017nih} is also a popular benchmark for multi-label classification tasks. It contains 112,120 frontal-view chest radiographs with 14 pathological observations annotated for each. We adopted the dataset partition in~\cite{zhou2023mrm}, which separates the entire dataset into a training/validation/test set by 7:1:2.

\textbf{RSNA Pneumonia}~\cite{shih2019rsna} defines a binary classification problem, where each chest radiograph is categorized as either pneumonia or normal. We adopt the official data split, where the training/validation/test set comprises 25,184/1,500/3,000 images, respectively.



\textbf{SIIM-ACR Pneumothorax Segmentation (SIIM)}\footnote{https://www.kaggle.com/c/siim-acr-pneumothorax-segmentation \label{kaggle:siim}} is a popular benchmark for pneumothorax disease segmentation. It contains over 120,000 frontal chest X-rays with precise manual annotations of pixel-level pneumothorax label. We adopt the data partition in MRM~\cite{zhou2023mrm}.

\begin{table*}[]
\centering
\caption{Datasets utilized in our experiments.}
\label{tab:dataset}
\resizebox{\textwidth}{!}{%
\begin{tabular}{@{}lccccccc@{}}
\toprule
\multirow{2}{*}{Datasets} &
  \multirow{2}{*}{Stage} &
  \multirow{2}{*}{Tasks} &
  \multicolumn{4}{c}{Number of Images in Splits} &
  \multirow{2}{*}{Annotations} \\ \cmidrule(lr){4-7}
 &
   &
   &
  Training &
  Validation &
  Test &
  Total &
   \\ \midrule
MIMIC-CXR~\cite{johnson2019mimic} &
  Training/Evaluation &
  R2Gen &
  222,758 &
  1,808 &
  3,269 &
  377,110 &
  227,835 Reports \\
\begin{tikzpicture}[baseline=(current bounding box.south)] \draw[-] (0,0.3) -- (0,0) -- (1,0); \end{tikzpicture} MS-CXR~\cite{boecking2022biovil} &
  Evaluation &
  Phrase Grounding &
  - &
  - &
  881 &
  881 &
  1,162 \\
CheXpert~\cite{irvin2019chexpert} &
  Evaluation &
  Classification &
  218,414 &
  5,000 &
  234 &
  224,316 &
  14 Multi-label Diseases \\
\begin{tikzpicture}[baseline=(current bounding box.south)] \draw[-] (0,0.3) -- (0,0) -- (1,0); \end{tikzpicture} Chexpert-5x200~\cite{huang2021gloria,zhou2023mrm} &
  Evaluation &
  Retrival &
  - &
  - &
  1,000 &
  1,000 &
  1,000 Reports \\
NIH ChestX-ray~\cite{wang2017nih} &
  Evaluation &
  Classification &
  78,484 &
  11,212 &
  22,424 &
  112,120 &
  14 Multi-label Diseases \\
RSNA Pneumonia~\cite{shih2019rsna} &
  Evaluation &
  Classification/Detection &
  25,184 &
  1,500 &
  3,000 &
  29,684 &
  Binary Classification for Pneumonia \\
SIIM-ACR Pneumothorax~\cite{siim-acr-pneumothorax-segmentation} &
  Evaluation &
  Classification/Segmentation &
  7,969 &
  1,898 &
  1,904 &
  11,771 &
  Binary Classification for Pneumothorax \\ \bottomrule
\end{tabular}%
}
\end{table*}

\subsection{Data Preprocessing}
\noindent \textbf{Image Preprocessing.} For our synergistic pretraining, we utilize all frontal and lateral images from the MIMIC-CXR dataset, randomly resizing and cropping each input image to $224 \times 224$ pixels. During training, the data augmentation only involves \texttt{RandomHoriz\-ontalFlip} and \texttt{RandomAffine}. The augmented images are subsequently normalized using a mean of $0.4978$ and a standard deviation of $0.2449$.

\noindent \textbf{Report Preprocessing.} For report generation, each frontal or lateral image is associated with a single report, which comprises either only the findings section or a combination of the findings and impressions sections. The extraction and cleaning functions are developed in accordance with R2Gen~\cite{chen2020r2gen}. Subsequently, each report is truncated to a length of 100 tokens following conventional methodologies~\cite{chen2020r2gen,chen2021r2gencmn,wang2023r2gengpt,gu2024comg}. 

\subsection{Synergistic Pre-training}
Our synergistic pre-training approach concurrently optimizes both visual and language components. For the visual decoder and the text encoder, we employ a 4 layer Vision Transformer~\cite{dosovitskiy2020vit} and a 12 layer Bert model~\cite{kenton2019bert} configured according to the previous works~\cite{huang2024maco}. For the language decoder, we utilize the GPT-2 model~\cite{radford2019gpt2}, configured according to the official settings\footnote{\url{https://huggingface.co/openai-community/gpt2}}. 

For the pre-trained image classifier used to predict potentially relevant pathologies during inference, we adopt a fine-tuned version of the visual encoder from SS-ACL, tailored by a linear classification head. During fine-tuning, we apply a linear probing protocol by freezing the parameters of the visual encoder and training only the linear head. This approach minimizes knowledge bias, reduces training costs, and maximizes the utility of the SS-ACL framework.

\subsection{Downstream Tasks}
Our implementation of downstream tasks encompasses the reproduction of fine-tuning procedures for classification~\cite{zhou2023mrm} and segmentation~\cite{huang2024maco}, as well as the execution of zero-shot methodologies for retrieval~\cite{huang2021gloria}, classification~\cite{huang2024maco}, and phrase grounding~\cite{huang2024maco}. The corresponding projects we followed are cited alongside each task.

\noindent \textbf{Fine-tuning Classification.} This task involves adding an additional linear head (with some studies also incorporating a normalization layer~\cite{zhou2023mrm,huang2024maco}) on top of the visual encoder and fine-tuning the combined model on the downstream datasets.

\noindent \textbf{Linear Probing Classification.} Unlike fine-tuning classification, linear probing freezes the visual encoder and only trains the linear head. This approach is more effective for evaluating the intrinsic discriminative capacity of the pre-trained visual features.

\noindent \textbf{Zero-shot Classification.} This task typically involves predicting the pathology class for radiographs by calculating the similarity between visual representations and templated language prompts~\cite{tiu2022chexzero,huang2024maco}. The design of the prompt template can significantly impact performance. To ensure fair and consistent comparisons, we utilize the empirical template provided by MaCo~\cite{huang2024maco}.

\noindent \textbf{Zero-shot Retrieval.} This task entails calculating the similarity between radiographs and their corresponding reports to retrieve image or text candidates, where the samples are unseen by the models.

\noindent \textbf{Zero-shot Phrase Grounding.} This task requires the explicit extraction of pixel-level relationships between radiographs and their corresponding ground-truth sentences. We employ DINO Attention~\cite{caron2021dino} to extract the inherent attention map from the last normalization layer of the Vision Transformer, along with the correlation between images and phrases, to compute the final attention map linking images to specific region descriptions.

\begin{figure}[]
    \centering
    \includegraphics[width=\linewidth]{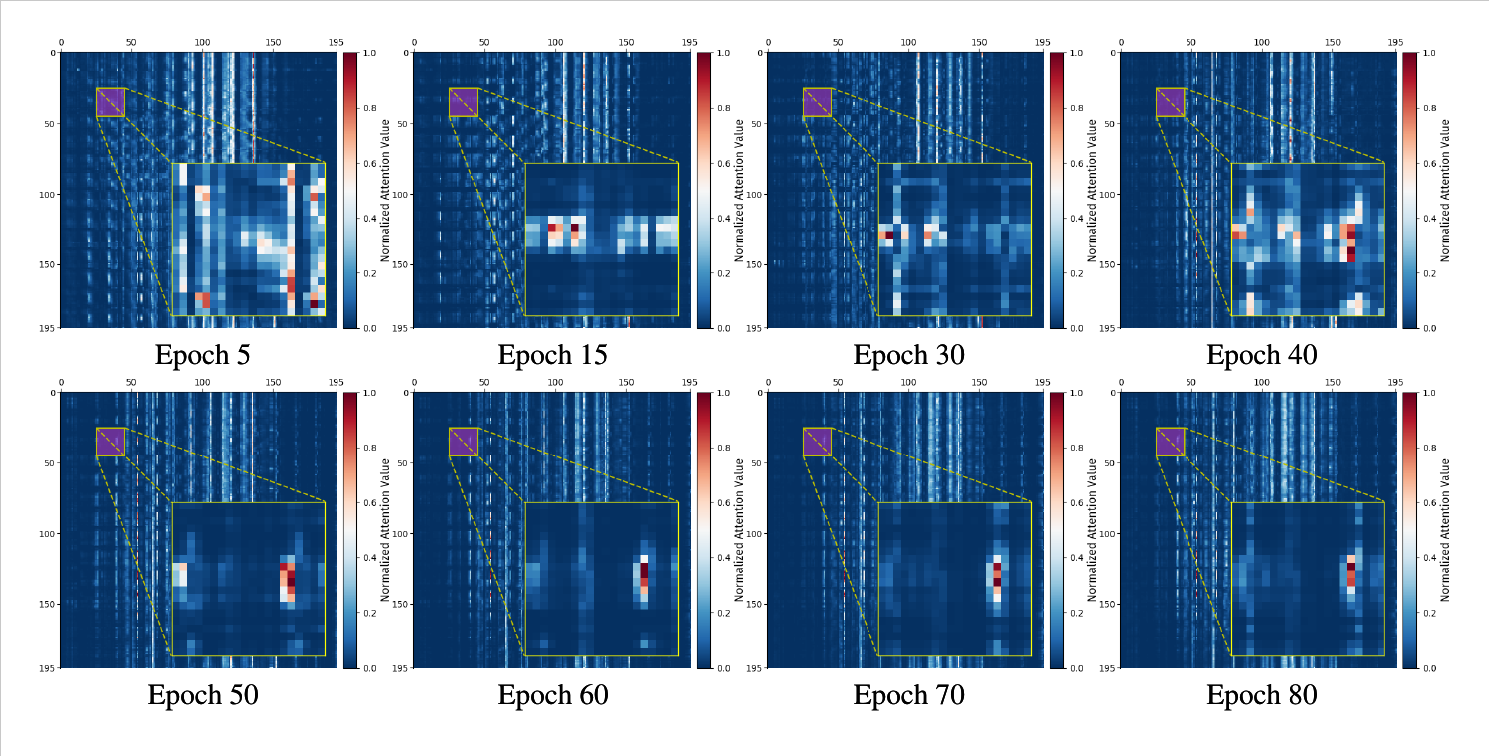}
    \caption{Comparison of inherent patch-wise attention maps generated by SS-ACL during training convergence. Rows correspond to query values in the Vision Transformer, while columns represent key values.}
    \label{fig:suppl-abs-vis}
\end{figure}

\section{Additional Qualitative Analysis}
\label{suppl:sec:ablation}

\noindent \textbf{Visual Discrimination.} Figure~\ref{fig:suppl-abs-vis} illustrates the normalized attention values between query and key representations derived from the visual encoder. Guided by anatomical prompts, the learned features exhibit clear discrimination during training convergence, demonstrating the effectiveness of the proposed fine-grained alignment strategy. However, as training progresses, the model tends to overemphasize the distinction between associated regions, leading to excessive sparsity in the feature space. This, in turn, hinders the clustering of anatomically consistent regions and negatively impacts generation performance.  

\begin{figure*}[]
    \centering
    \includegraphics[width=\linewidth]{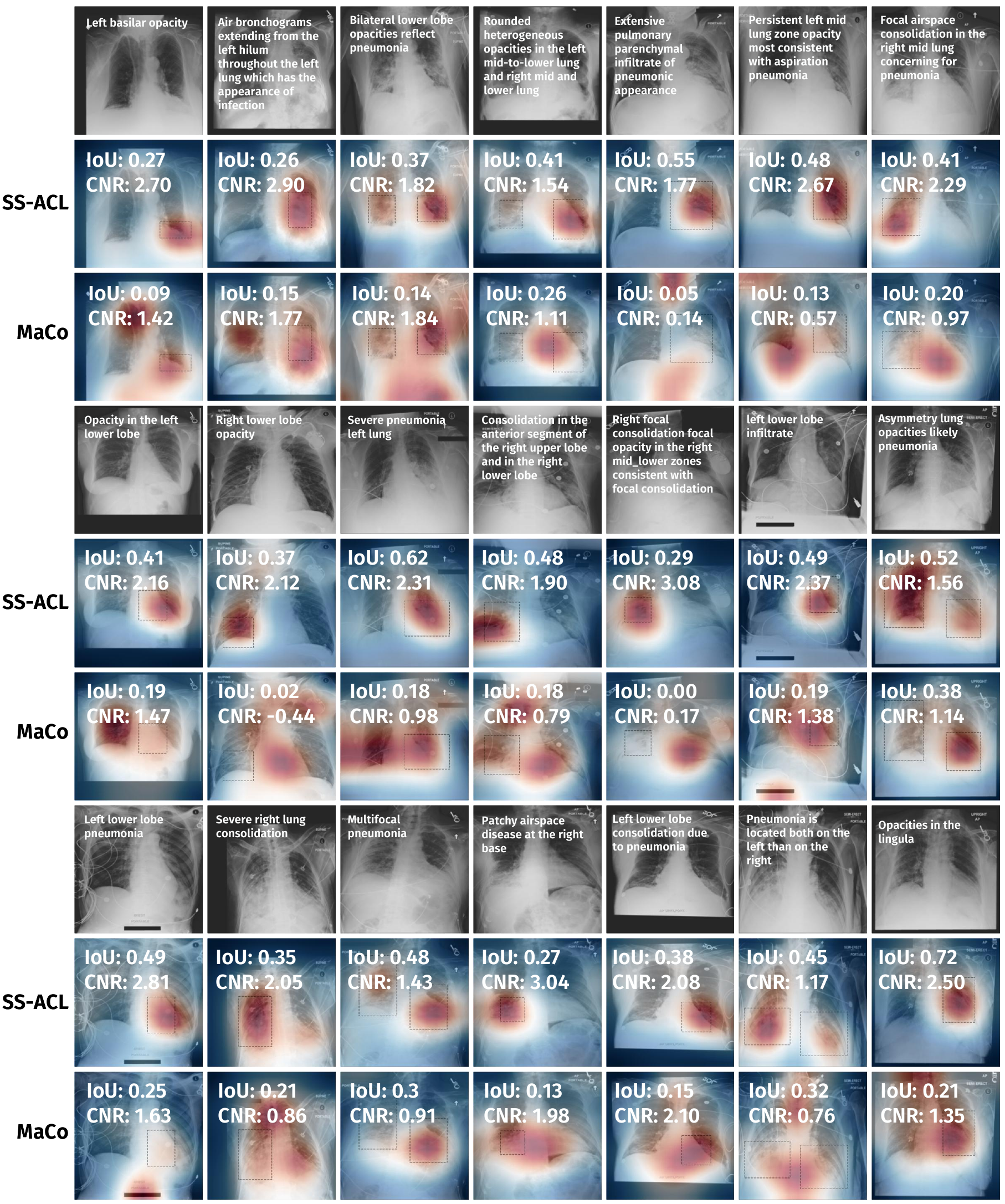}
    \caption{Qualitative results of phrase grounding on MS-CXR. The ground-truth bounding boxes are drawn with the corresponding region descriptions.}
    \label{fig:suppl-vis-pg}
\end{figure*}

\noindent \textbf{Phrase Grounding.} In Figure~\ref{fig:suppl-vis-pg}, we present additional samples from the MS-CXR dataset to illustrate the inherent interpretability of SS-ACL in identifying relevant regions based on fragments of radiological descriptions. Typically, MRG models struggle to extract meaningful attention maps for specific phrases or sentences, as their visual representations are utilized solely for text generation without equivalent interactions or alignment between language and vision. In contrast, SS-ACL not only delivers high-quality language generation but also facilitates effective interactions from language to vision. This synergistic property underscores the significant potential of automatic diagnosis models to advance the development of explainable artificial intelligence in medical diagnostics.

\begin{figure*}[]
    \centering
    \includegraphics[width=0.93\linewidth]{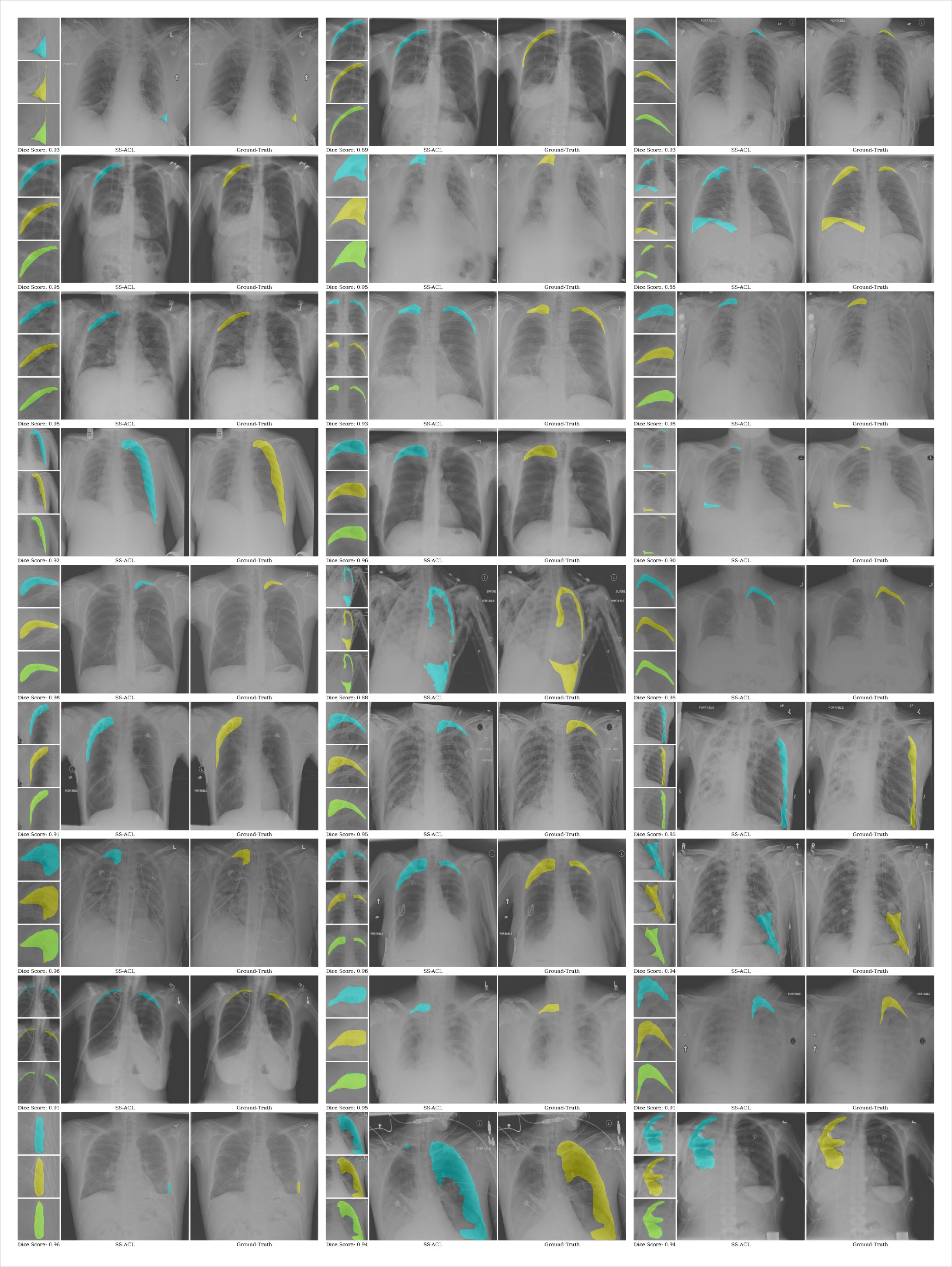}
    \caption{Visualization results of segmentation on SIIM. The ground-truth segmentation masks are annotated by yellow, the predicted masks by SS-ACL are in blue, and the green masks represent the interaction between them.}
    \label{fig:suppl-vis-seg1}
\end{figure*}

\noindent \textbf{Pneumonia Segmentation.} Figure~\ref{fig:suppl-vis-seg1} showcases the visualization results of pneumonia segmentation on the SIIM dataset~\cite{siim-acr-pneumothorax-segmentation}. SS-ACL demonstrates exceptional generalization capabilities by accurately identifying even minute pneumonia regions, highlighting its robustness in clinical applications.

\begin{figure*}[]
    \centering
    \includegraphics[width=\linewidth]{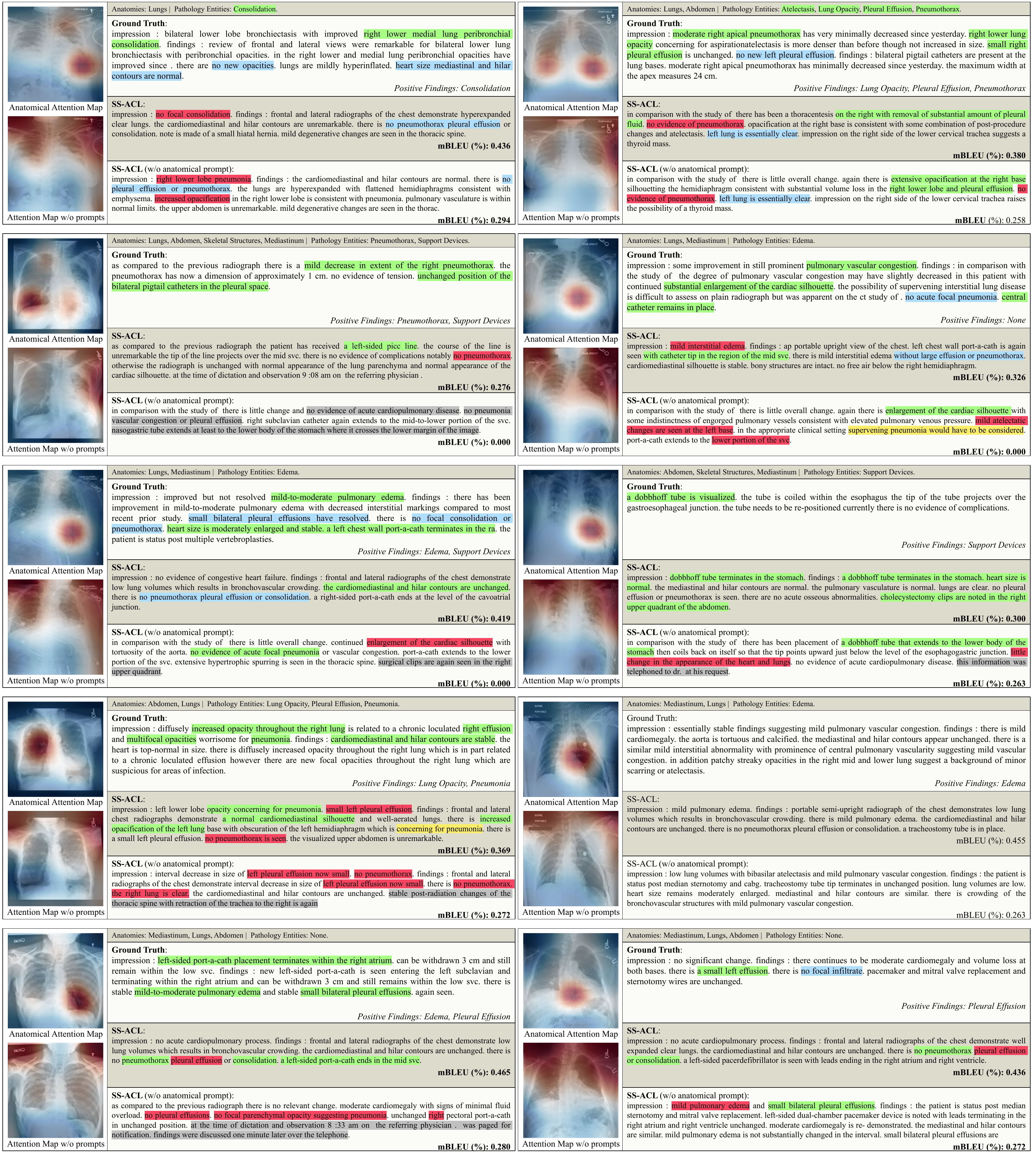}
    \caption{Additional visualization of report generation on MIMIC-CXR. The green highlights indicate detected positive abnormalities, yellow highlights denote uncertain pathologies, red marks indicate incorrect estimations, and gray highlights represent unrelated descriptions.}
    \label{fig:suppl-vis-rrg}
\end{figure*}

\noindent \textbf{Report Generation with Anatomical Prompts.} In Figure~\ref{fig:suppl-vis-rrg}, we display additional report generation results to evaluate the effectiveness of our proposed anatomical prompting strategy. The results indicate that precise prompts significantly enhance the generation of accurate and relevant findings. Specifically, when different prompts guide the model to focus on distinct pathologies or anatomical areas, SS-ACL adeptly adjusts its focus in accordance with the prompt instructions. This demonstrates SS-ACL's strong flexibility and adaptability in clinical practice, particularly when physicians or radiologists provide detailed and targeted instructions.

\section{Discussion and Future Work}
\label{suppl:sec:discussion}

SS-ACL's exceptional synergistic performance has been consistently highlighted and validated through comprehensive experiments and systematic analysis. Nonetheless, as the first endeavor toward such an extensive objective, SS-ACL presents certain limitations within its current framework.

\noindent \textbf{Prompting Mechanism and Dependency.} The primary limitation concerns the inherent mechanism of the proposed prompting strategy, which significantly enhances report generation. A critical issue to address is that while SS-ACL can automatically generate prompts without manual intervention—thereby ensuring the fairness of all experiments—the prompts provided to the language generator may inadvertently introduce prior information. Specifically, the pathologies mentioned in the ground-truth reports might not accurately reflect the original clinical data but instead amalgamate the intrinsic relationships between abnormalities. As a result, the model may predict final outcomes based on these priors rather than the authentic signals derived from the visual representations. In future work, we aim to explore the underlying mechanisms of the prompting policy and further refine the effectiveness of visual features to enhance the reliability of prompt-guided radiology report generation.

\noindent \textbf{Integration with Large Language Models (LLMs).} Currently, SS-ACL operates as a full-parameter-training model, which may be computationally expensive for integrating large language models. Developing effective adapter designs and leveraging transfer learning for SS-ACL constitute additional avenues for future research. We believe that the demonstrated potential of LLMs, widely recognized in the natural image domain, holds substantial promise for the medical field and aligns with our objectives.


\end{document}